\documentclass[]{interact}

\usepackage{epstopdf}%
\usepackage[caption=false]{subfig}%
\usepackage{pdfpages}
\usepackage[linesnumbered,ruled,vlined]{algorithm2e}

\usepackage{amsmath,amssymb}

\usepackage{todonotes}
\usepackage{url}

\usepackage[authoryear]{natbib}%
\makeatletter%
\def\NAT@def@citea{\def\@citea{\NAT@separator}}%
\makeatother%

\theoremstyle{plain}%

\theoremstyle{definition}

\theoremstyle{remark}

\title{DBOT: Artificial Intelligence for Systematic Long-Term Investing}

\author{
\name{Vasant Dhar\thanks{Email: vd1@stern.nyu.edu} and Jo\~{a}o Sedoc\thanks{Email: jsedoc@stern.nyu.edu}}
\affil{New York University}
}

\begin{document}

\maketitle

\begin{abstract}
Long-term investing was previously seen as requiring human judgment. With the advent of generative artificial intelligence (AI) systems, automated systematic long-term investing is now feasible. In this paper, we present DBOT, a system whose goal is to reason about valuation like Aswath Damodaran, who is a unique expert in the investment arena in terms of having published thousands of valuations on companies in addition to his numerous writings on the topic, which provide ready training data for an AI system. DBOT can value any publicly traded company. DBOT can also be back-tested, making its behavior and performance amenable to scientific inquiry. We compare DBOT to its analytic parent, Damodaran, and highlight the research challenges involved in raising its current capability to that of Damodaran’s. Finally, we examine the implications of DBOT-like AI agents for the financial industry, especially how they will impact the role of human analysts in valuation.
\end{abstract}

\section{Introduction}
\label{sec:intro}
We present an AI system called DBOT that makes recommendations for long-term investing by reasoning like an established human valuation expert, Aswath Damodaran, who is often referred to as the ``Dean of valuation'' in the financial industry. DBOT can value any public traded company on the basis of Damodaran's analysis, and generates a report to support its position in an attempt to mimic its analytic parent. Until recently, such capabilities of analytic twins for financial valuation were not feasible. However, with advances in large language models (LLMs) and generative artificial intelligence (GenAI), it has become possible to conduct valuations that marry numbers and reasoning to generate credible valuations that can be used for long-term investing. The implications for automation and support of various parts of the valuation exercise are profound. 

In this paper, we provide a method for creating a digital analytic twin, DBOT, which is designed to mimic the investment analysis of individual companies by Damodaran. Since DBOT can value every company in an index such as the S\&P500, it also provide an analysis in a macro sense, for example, by valuing the S\&P500 market index relative to the valuation of its individual components. 

From the perspective of generative AI, DBOT presents a multitude of challenges. First and foremost, LLMs must be able to reason over financial texts, charts, tables, and spreadsheets. Furthermore, DBOT requires the AI system to follow Damodaran's specific strategy rather than the generic fundamental reasoning that large language models (LLMs) have learned from the Internet. To the best of our knowledge, this is the first attempt to create a digital analytic twin of an investment expert that ``thinks'' like the expert, based on the available training data of the expert. Fortunately, Damodaran has made all his analyses public and is the only expert in finance who has published thousands of company reviews on the Internet. This provides a rich set of training data for creating and evaluating the digital analytic twin

\section{Valuation and Systematic Investing}
\label{sec:valuation}
Historically, the financial services industry has been one of the leading sectors to adopt technology, beginning with the automation of trading marketplaces and transaction processing~\citep{steiner2012automate}. The New York Stock Exchange (NYSE) introduced electronic order matching in the early seventies, and depository trust institutions automated the post-trade process of settlement and record keeping~\citep{weiss2006after}.

Faster moving markets, increasing liquidity, and the availability of data created by automation, moved computers upstream, replacing traditionally human roles such as market making with algorithms. Pre-trade analysis, which had traditionally been entirely human-based, also became increasingly automated, leading to fully automated decision-making algorithms that sent orders directly to the marketplace. 

Over the last two decades, short-term trading, which involves holding periods of days to weeks, has become increasingly automated as portfolio managers searched for new sources of ``alpha'' on the basis of short-term price movements. Alpha refers to the ability of an investment strategy to generate returns that are uncorrelated to a reference benchmark (e.g., a market index). 

A class of algorithms that were based on historical price data, loosely referred to as statistical arbitrage or ``Stat Arb'' \citep{grinold2000active} emerged during the eighties for systematic long/short investing based on linear regression models. These were followed by machine-learning-based AI algorithms~\citep{dhar2000discovering}, which search for latent factors that drive returns. Such algorithms typically learn trading rules that are described in terms of latent factors and are now commonplace in the investment world~\citep{de2018advances}.

Short-term trading algorithms typically have little to do with the fundamentals of companies, which change slowly. Instead, they are driven primarily by prices and secondarily, from periodic data releases from business and government agencies. Such algorithms learn the trading rules from the data.

The classic book by \citet{graham1934security}, called Security Analysis, is the first attempt to systemize valuation based on fundamental data. It has served as a bedrock for investing in stocks and fixed income instruments for the last century. The legendary investor Warren Buffett was a student of Graham and Dodd in the early fifties in New York, and acknowledges their role in his intellectual development in the area of value-based investing. Such an analysis requires estimating the expected cash flows of a company over time and discounting these to a single value: the price of a stock.

So far, AI methods have not been feasible for long-term investing due to the difficulty of creating a suitable knowledge base for investing. This is because there has been no specifiable theory to create the relevant knowledge for the AI, nor has there been sufficient training data from which a machine learning algorithm could learn the relevant knowledge for valuation. In other words, there is no Graham and Dodd algorithm for value investing, nor has there been sufficient training data to learn such an algorithm. Rather, the workhorse for systematic quantitative investing in the industry over the last thirty years has been the factor model, which attempts to explain a stock's return in terms of a multitude of factors such as the return of the overall market, the stock’s industry membership, and other firm-specific factors such as its size, fundamentals, and momentum.

Factor models have their roots in the single factor capital asset pricing model (CAPM) proposed by~\citet{sharpe1964capital}, which breaks down a stock's return into two parts: that accounted for by the market and the part that is idiosyncratic to the stock. Over the years, dozens of factor models have emerged that expand on CAPM. For example, the Fama-French model~\citep{fama1993common} introduces two factors in addition to the market in order to explain the returns of individual US stocks: the size of a company based on market capitalization, and ``value,'' based on the price-to-book ratio. Researchers have also proposed other ways of measuring the value factor, such as earnings, cash flows and sales~\citep{asness2013devil}.

The literature on factor models is extensive~\citep{giglio2022factor}, and includes non-fundamental factors such as momentum~\citep{jegadeesh1993returns,carhart1997persistence}. Some approaches have derived momentum and other latent factors using a principal components analysis (PCA) applied to returns data~\citep{chamberlain1982arbitrage}. Factor models of various flavors are commonly used in the financial industry, which combine some mix of fundamental and technical factors such as momentum and volatility. Such models can be backtested, that is, tested on historical data. This makes them systematic, which is attractive to portfolio managers who can tilt their factor exposures to suit their desired investment objectives.

Comprehensive critiques of multi-factor models have been explored in various research papers that have examined their theoretical and empirical limitations~\citep{harvey2016and}. An analysis by \citet{kelly2019characteristics} details the limitations of multi-factor models to explain the cross-sectional variability in stock returns and notes how they struggle to account for real-world anomalies. Their takeaway is that static factor models do not fully capture the complexities of market dynamics and context, which require the consideration of other time-varying factors that are difficult to boil down into traditional static risk factors. As a simple example of one such nuance, for example, a consideration of the competitive dynamics within an industry that might be revealing itself in current news requires a deeper reasoning process based on the current context than what traditional factor models can handle. Ultimately, factor models only use surface-level characteristics of companies and assets; making them akin to machine-learning AI algorithms. Humans often make adjustments, based on context.

In contrast, we posit that this type of contextual reasoning ability is an aspect of what is unique about valuation experts such as Aswath Damodaran and Warren Buffett. They attempt to integrate the numbers from financial statements, news, political landscape and other information sources alongside a narrative into a coherent story that captures the present context and valuation. 
Our objective is to systematize Damodaran’s contextual reasoning ability into an algorithm. This makes long-term investing systematic and backtestable for the first time.

\begin{figure}[h]
    \centering
    \includegraphics[width=0.9\linewidth]{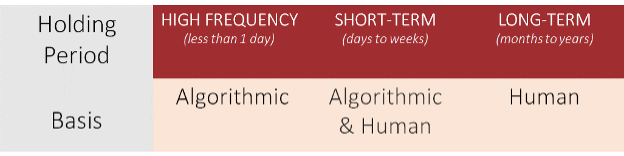}
    \caption{Basis for Decision-Making By Holding Period}
    \label{fig:holding-period}
\end{figure}

Figure~\ref{fig:holding-period} shows how algorithms have been applied to date across the investment landscape. We categorize high-frequency trading as involving holding periods of less than a day, short-term trading in terms of holding period to days to weeks, and long-term investing in terms of holding periods of months to years. 

In the investment landscape of Figure~\ref{fig:holding-period}, market making and high frequency trading were the first to give way to algorithms across all markets. Current   estimates are that high-frequency trading is almost completely automated, and automated short-term trading is accounting for an increasing part of trading volume. Until now, AI-based long-term systematic investing has not been feasible for reasons we have described above, namely, the unavailability programmable  human expertise or  training data for valuation. In other words, it hasn’t been feasible to translate long-term investing into an algorithm either through top-down specification or bottom-up machine learning.

The emergence of LLMs and GenAI has changed this with the near advent of “general intelligence,” which is a blend of common sense and domain knowledge, as represented in the training data (e.g., language on the Internet). Serendipitously, by learning how to predict the next word in a context very well, LLMs have learned how to simulate human reasoning to the point where we often cannot tell the difference between human and machine generated outputs.

The challenge is how to integrate AI’s general intelligence with specialized knowledge that incorporates the depth of deep human thinking of human experts like Damodaran. Such an AI system should be backtestable, and provide a method for systematic long-term investing that can be analyzed scientifically. 

\section{Artificial Intelligence and the Damodaran Bot}

The field of Artificial Intelligence has gone through three paradigm shifts in the last sixty years, where each shift addressed a major bottleneck encountered by the existing paradigm~\citep{dhar2024paradigm}. Early AI research during the late 50s and 60s was dominated by game playing search algorithms~\citep{samuel1959some} that led to novel ways for searching various kinds of graph structures. But this type of mechanical search provided limited insight into intelligence, where real-world knowledge seemed to play a major role in solving problems, such as in medical diagnosis and planning. Expert Systems provided a way forward, by representing domain expertise and intuition in the form of explicit rules and relationships that could be invoked by an inference mechanism. But this knowledge was hard to extract and the systems were hard to create and maintain. This was a major bottleneck in getting knowledge into the machine in a scalable way.

The first major paradigm shift was from the use of expertise to the use of data, which started becoming plentiful as information technology matured. The second paradigm shift, towards deep learning, moved intelligence upstream, closer to the source of the data, where machines could perceive their environment directly through vision, sound, and language instead of requiring us to craft and curate features from data for them. For example, you didn't have to featurize an image for the algorithm, but could feed the pixels from the image directly as input into an algorithm. Language-based learning also became feasible from the gobs of data on the Internet, requiring minimal curation and filtering of the inputs. Machine-based perception opened the floodgates for autonomous machine learning.

The latest shift, to general intelligence, is about machines that can learn from any kind of data and reason about any topic using any data type (except for smell and touch at the moment) right out of the box (e.g., ChatGPT).  Pre-trained large language models (LLMs) on which applications such as ChatGPT are built, have encapsulated the collective expression of humanity on the Internet, and have become capable of reasoning about nearly any situation in natural language.

Unsurprisingly, ChatGPT already knows a lot about Aswath Damodaran and other valuation gurus like Warren Buffett. That's what general intelligence is all about – the machine knows quite a bit about everything. Which raises the obvious question, whether AI systems such as ChatGPT are good enough to make long-term investment recommendations right out of the box. 

\citet{fairhurst2025how} have found that while ChatGPT-4 can answer basic financial questions, the answers are still frequently incorrect and don’t reflect sufficient knowledge about Finance.

Our experience is similar, that AI systems are not sufficiently trustable for valuation, and it's not only because of the lack of knowledge but also their lack of stability. For example, in early November of 2024, we asked ChatGPT for a yes or no answer to whether it would buy any of the following with a one-year investment horizon in mind: the S\&P500, Microsoft, Google, BYD, and Nvidia. It said ``yes'' to all of them except Nvidia, for which it explained ``Long-term growth prospects may remain strong, but short-term market corrections, competition, and changing market dynamics could affect returns over your investment horizon.''

This was a reasonable response, but minutes later we posed essentially the same question, but the query didn’t include a required yes or no answer. This time, it listed the pros and cons of each investment and stopped without answering the question. When pressed for an explicit yes or no, it reversed its decision on Nvidia and BYD even though the question was, in effect, exactly the same.  Forcing it into a two-stage chain of thought reasoning~\citep{wei2022chain} that made it decompose the problem explicitly into the pros and cons of each investment resulted in a different answer. It’s hard to trust this kind of instability of the AI system, where its decisions flip in response to minuscule syntactic prompt variations. Importantly, it also makes it difficult to backtest such a system, given the near infinite ways to repose the same question on top of the already stochastic generative process.

Given the instability of direct LLM valuation, we explored fine-tuning the LLM along the lines of recent work that has fine-tuned LLMs for general purpose financial applications~\cite[][et alia]{wu2023bloomberggpt,xie2023pixiu,zhang2023instruct,wang2023finvis,bhatia-etal-2024-fintral}. We fine-tuned GPT-4o on Damodaran's ``Musings on Markets.'' However, this fine-tuned LLM produced reports in the linguistic style of Damodaran, but failed to capture his analysis and thus lacked credible valuations for companies. The valuations were based on internal model parameters instead of an external valuation tool, resulting in its reports sounding like a generic analyst. 

Recently, Google's Gemini~\citep{GoogleGeminiDeepResearch}, OpenAI's o3-mini~\citep{openai2025deepresearch}, and other models included a ``deepresearch'' mode. This so-called deepresearch is defined by 1) a planning mode, in order to create a multi-step process, 2) a search mode, to look up relevant websites and internet resources, 3) a reasoning mode, to reason over all of the sources and align them with the rest of the research plan, and finally 4) a writing mode, to synthesize the information from the sources. We prompted both Gemini-1.5 and o3-mini deep research models with the requirement to analyze in the style of Damodaran and found that while the reports were superior to fine-tuning there were two central flaws. Not surprisingly, the first of these was that the framing for valuation was largely dominated by previous valuations from other sources. This meant that it was largely a general amalgamation of news stories on the Internet. The second major flaw was a very superficial consideration of fundamentals such as a free cash flow analysis of the company being valued. Including such data into the planning phase did little to deepen the analysis, let alone produce anything comparable to Damodaran's valuations.

Interestingly, however, what our simple example showed was the benefit of breaking down the problem into successively deeper levels, which leads to more explicit reasoning and transparency. Thus, the approach we took is to break down Damodaran’s valuation process into several specialized ``agents,'' each designed to address a limited part of the valuation exercise. Our approach is agnostic to the choice of LLMs, which are essentially pluggable modules into DBOT. As they improve, so should DBOT.

DBOT produces a report through its interaction with its specialized agents. %
Such multi-agent systems involving LLMs have recently gained popularity for addressing complex tasks such as code design~\citep{hong2024metagpt}, scientific inquiry~\citep{lu2024ai}, and financial analysis~\citep{zhou2024finrobot}. A decomposition into functional ``units'' accomplishes several critical objectives: it allows each agent to focus on a particular task, enabling clearer design of the unit; it simplifies the testing of individual components through ``unit testing''; and it permits the selection of generative AI models with capabilities specifically suited to certain subtasks, such as dedicated coding agents or multimodal agents for tasks requiring visual processing. FinRobot~\citep{zhou2024finrobot} is an AI financial analyst designed as a multiagent system.  But FinRobot inherits the same deficiencies as generic LLMs, of producing reports and valuation which are too generic and surface level to be trustable. Nonetheless, such systems can potentially aid analyst productivity.

In summary, prior methods that rely on the generic knowledge available on the Internet to learn about valuation are not sufficiently grounded to be trustable. They are not based on a methodological approach that is vetted in expertise, such as Damodaran’s. This is one of the main unique contributions of DBOT.

Our main question--which could have profound implications for the investing profession--is ``can LLMs replicate Damodaran's analytic perspective on investments?'' We further break this down into the following types of questions:  Can DBOT reason over Damodaran's free cash flow valuation model? 
Can we increase the stability and factual correctness of DBOT? Can DBOT generate a blog post in the style of Damodaran? In the course of building and testing DBOT, however, a challenging question we have encountered, and one that challenges the current state of AI is the following: can AI creatively find good ``framing questions'' for valuation that mimic Damodaran?

Assuming that the questions we raise are not theoretically insurmountable, we should expect systems such as DBOT to get much better over time and to be incorporated into industry practice, which raises the following questions: Will we need human analysts anymore, or will AI do a comparable job or even better? Alternatively, how will AI machines such as DBOT change the role of human analysts in the industry? 

\section{DBOT Architecture}

\subsection{Agents}

The roots of multi-agent systems go back to the 1970s when the AI pioneer and Turing award winner Alan Kay modeled intelligence as being distributed among objects that communicated through ``messages'' telling them what to do. Kay worked for Atari, so video games provided a use case for his agent-based model. Messages were precisely coded, such as ``turn $<$object$>$ 180 vertical,'' which means apply the ``turn method'' to the object of interest with a 180-degree rotation along its vertical axis. These days, LLM-powered AI agents can process more complex messages written in human language (in our case English), such as ``summarize the current news sentiment related to BYD whenever there is news.'' This kind of intelligence makes current-day AI agents powerful in the sense that they can figure out how to do something in a given context by merely being told what to do. In other words, agents now can be programmed in natural language.

\begin{figure}
    \centering
    \includegraphics[width=0.8\linewidth]{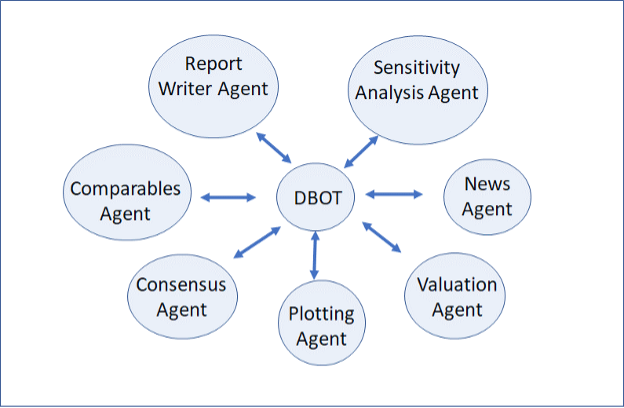}
    \caption{The DBOT Architecture}
    \label{fig:dbot-arch}
\end{figure}

In DBOT, a set of independent agents are responsible for handling specific parts of the valuation exercise. They are orchestrated by a supervisor agent, who composes the valuation report for any publicly traded company globally. 
Figure~\ref{fig:dbot-arch} shows the agent architecture for DBOT. 

Damodaran’s analytic strategy is defined by ``stories to numbers.'' This is the fundamental tenant of his analytic framework. The ``numbers'' are then processed through his valuation model called the ``Ginzu.'' The supervisor, valuation, sensitivity, comparables, and consensus agents all have access to the valuation model to change the inputs. The report writer agent only has ``read access'' to the inputs and their associated valuation, and cannot change the inputs. 

All models other than the News agent are provided with Damodaran’s explanation of his methodology which he explains in a video that describes the ``Ginzu'' model for valuation.

The quantitative valuation agent calculates the value of a company according to Damodaran’s methodology. In addition to certain macro factors such as interest rates, the valuation agent uses four company-specific ``value drivers'' to value a company: sales growth, operating margin, the cost of capital, and reinvestment efficiency. The last of these is measured as the ratio of sales to capital, which indicates how much in sales a company generates per dollar of capital. The valuation agent obtains all its inputs from external financial databases containing the latest balance sheet, income statement, and cash flow statements. Based on these inputs and relevant macroeconomic data. It outputs a valuation for the company. 

The Consensus and Comparables agents obtain consensus estimates of analysts, as well as the companies they regard as being comparable in some way, typically those belonging to the same industry. These agents access the relevant consensus estimates and expectations of analysts from external databases. 

The Sensitivity Analysis agent considers alternative scenarios to vary the value drivers to generate the appropriate valuations. For example, if DBOT wants to modify its valuation to bring it closer to that  of the market, it would ask the sensitivity analysis agent to adjust the value drivers accordingly. Unlike other agents, the Sensitivity Analysis agent will ``reason'' over the spreadsheet by iteratively observing changes to the valuation with inputs. The valuation agent will often return tables of such data to the report writer. 

The News agent scours various news sources for the latest information on the company, the industry, and any relevant political activity. Its sources include the Wall Street Journal, the Financial Times, and a number of other media outlets. It interprets the news and returns relevant information that may result in further adjustments to the value drivers. Unlike a standard summarization agent, the News agent does a task-based summarization which it only extracts information that is specifically important for fundamental valuation. The News agent emulates human analysis by ``reading'' news articles in three phases 1) headline only, 2) headline and the first paragraph, and 3) full article. At each point it reflects on the importance of such news for valuation. After collecting the news from multiple sources, is assimilates it into a final summary for DBOT. The News agent also includes images and figures from various news sources into the summary if they are deemed interesting.

Once the calculation and reasoning agents are done with their analysis, two additional agents create the final report. The first, the Plotting agent, is designed to produce the visuals to support the narrative. The second is responsible for writing up the final report. It consists of two sub-agents: a Writer and a Critic. The responsibility of the Critic is to ensure that there are no loose ends in the report, the data are verified, all sources reported, and the report is of the right length. The Plotting agent creates the visuals corresponding to the narrative.

\subsection{The Algorithm}

The algorithm is presented in Appendix~\ref{app:algorithm}. The first eight bullets specify the required inputs and functions used by DBOT. Inputs include the company name $C_n$, its ticker symbol $C_t$, and data $D_C$ about its fundamentals from its financial statements. The symbol $I$ represents the inputs for Damodaran's Ginzu spreadsheet such as the fundamentals and value drivers, which are used by Damodaran’s valuation model represented by the function $f^{fcf}$ to calculate the value of a company designated by the symbol $v$.  The news function s takes the ticker symbol $C_t$ as input and returns the latest news for the company in the form of text. Finally, $f^{llm}$ designates the LLM model – such as ChatGPT – used by the application, which takes text and tabular information (and possibly images) as input, and returns text as output in the usual way as we interact with LLMs. Although we used OpenAI’s GPT-4o series models, it is possible to plug in other models such as Claude Opus, which may be better for certain tasks. (e.g., Report writer). This modular architecture of DBOT allows us to swap in more powerful agents as the technology improves. 

The six specialized functions use the LLM ($f^{llm}$) to produce their outputs, which are based on a prompt P to each agent and some additional relevant data. For example, the input to the function $f^{vd}$ is $I$, the spreadsheet which is based on fundamentals, along with a prompt “Based on the current spreadsheet, calculate the value drivers for the stock.” The function fDnews returns the relevant news which is incorporated into $I$, and  $f^{valuation}$ produces updates to the spreadsheet and value drivers that are used by the valuation agent. The News agent $f^{news\_summary}$  summarizes the news, and a plotting function, $f^{plot}$ creates the required visuals from the spreadsheet. The function $f^{report}$ generates the final report.

How do the agents interact to produce the final report? DBOT begins with a waterfall model where each agent is called in turn. This includes an initial valuation, which is followed by a sensitivity analysis, analyst consensus estimates, and an analysis of companies deemed comparable by analysts. While the ordering of this waterfall may initially seem arbitrary, we followed the logic of Damodaran’s analysis. He starts with an initial framing question which informs his initial decision, and questions his numbers based on the market and changes in the initial assumptions for the numbers. 

After DBOT has inputs from all its agents, it switches to a more contextual control that depends on two considerations. The first is the difference between its estimated value and that of the market. The second consideration is what the various agents have passed back to it in the previous round. Based on these, DBOT pursues what it deems to be most interesting avenue to explain the difference between the estimated and market valuations. This iterative process is specified in the ``while'' loop of the algorithm in Appendix 1. 

The iteration step is complete when DBOT’s estimate becomes stable. It then calls the report writer to create the final report with the appropriate text, charts, and tables. 

It is worth bringing attention to the fact that after the initial waterfall call to the agents, DBOT's attention is driven by context. This is one of the gifts and risks of the general intelligence paradigm – the machine has sufficient intelligence that a prompt such as “call the most appropriate agent” leaves it up to the machine to decide to which agent it passes control. In effect, we abdicate control of attention to the LLM’s meta-“attention mechanism,” thereby letting the layer of general intelligence make the decision about what to do next. This is another concept borrowed from multi-agent systems where we allow for the search to be directed depending on context. In effect, control is not specified in the application, but left to the general intelligence of the LLM. Exploring potentially better control mechanisms is an interesting research challenge.

Appendix~\ref{app:byd-report} lists the report DBOT created for BYD on November 4, 2024. It is typical of the kind of report it produces for any publicly traded company. The entire valuation process takes a few minutes to run for a company. This makes it feasible to evaluate all components of an index such as the S\&P500 in a day.

\section{Example: BYD Valuation}
DBOT valued BYD several times during October and November of 2024. Appendix 2 contains one such report that was generated on November 4, 2024, valuing BYD at \$420 per share. DBOT named the report “BYD in 2024: Riding the EV Wave or Struggling Through the Competitive Storm?” 

The title reflects the contents of the report, which balances the growth prospects in a growing industry against the increasing competition in the space from Tesla and other Chinese EV makers. Some of the titles it picked going back to mid-October of 2024 were equally interesting:
\begin{itemize}
    \item 	BYD: The Rise of the Dragon in the Electric Vehicle Market
    \item 	BYD: Navigating the Electric Vehicle Landscape with Strategic Precision
    \item 	BYD in 2024: Racing Forward or Running on Empty?
    \item 	BYD: Navigating Growth in a Shifting EV Landscape.
    \item 	BYD: A Comprehensive Analysis of Growth, Market Position, and Future Prospects
\end{itemize}

DBOT's more bullish titles in the first two reports from mid-October of 2024 talked up BYD’s battery technology, and its position as an established company with a global footprint and a demonstrated ability to scale. In the later reports closer to the US Presidential election, DBOT was more influenced by looming threats of tariffs and trade wars in the news. Perhaps the AI was picking up on Trump’s potential victory and his threats of steep tariffs on Chinese goods, and hence the tone of caution.

It is worth noting that despite variations in the titles and reports from one week to another due to the news, DBOT's valuations were was remarkably stable over the month and higher than its market price. This is a buy recommendation. As of this writing, BYD stock is up considerably since its recommendation.

Importantly, unlike ChatGPT, DBOT is not prone to changing its decisions based on small variations in prompts or minor changes in the business environment. This means that its performance can be backtested, in the same way as traditional quantitative trading strategies. This property is essential for determining trust. 

More generally, the stability implies that DBOT’s properties can be analyzed systematically and compared to its analytic parent. For example, Damodaran admits that one of the properties of value investing is that it tends to sell the big winners too early. Nvidia and Facebook are two prime examples where Damodaran sold too early. This isn’t surprising since value investing ignores factors such as psychology and momentum, that can drive asset prices for significant periods of time. Is DBOT similarly biased? It is testable.

There are several key points from the report that are worth highlighting, each of which influences one or more of DBOT’s value drivers, and its final value estimate.

The report shows BYD’s deliveries of EVs increasing steadily for the last three years except for a blip in early 2024 due to tariffs from the EU and disputes with foreign regulators. DBOT breaks down projected sales across different parts of the world, estimating total global sales rising from roughly 8 million EV sales today to 60 million by 2040.

With this assumption, BYD compares favorably to its competitors in terms of an “enterprise value to earnings” ratio (specifically, DBOT calculated the Terminal Enterprise Value to EBITDA (Earnings before Interest, Taxes, Depreciation, and Amortization)) ratio. BYD’s ratio, at 8.8, is almost half of Tesla’s and lower than that of the Chinese EV makers Nio, XPeng and Li Auto. All things being equal, it is cheaper than its competitors.

DBOT highlights the values of the four value drivers: revenue growth of 10\% for the next year, tapering down to 7\% in years 2-10, and 4.4\% thereafter; a target operating margin of 5.9\% for the most recent year, dropping to 5\% next year due to larger expenditures, and increasing to 6.7\% in years 2-10, and 7\% thereafter;  reinvestment efficiency is 1.2 for the next five years, and improves to 1.6 in years 6-10 due to lower subsequent capital investments. 
DBOT concludes with a sensitivity analysis shown in a table, in which we see that the valuations for the range of scenarios considered is between \$417 and \$468 per share. Its valuation of \$420 is at the lower end of this range.

At the end of the report, DBOT notes that the competitive landscape and regulation in each part of the world is different, so a “one size fits all” strategy won’t work. Rather, BYD will need to be adaptive to the specifics of each market.

An important question is whether you can tell the difference between DBOT’s report and one generated by a professional. Most humans cannot tell, so perhaps the more important question is whether the DBOT truly “understands” anything it has written or just appears to do so. 

Regardless of the answer, the  more important question is how good is DBOT? This is a harder question to answer, and for now, we can only compare it to its analytic parent. Although the machine performs a complex cognitive task, it does not have consciousness or the kind of subjective understanding of the world that humans use to do such tasks. This should make us examine its outputs carefully. While we found no errors in DBOT's BYD report, knowing that it is machine-generated should make us analyze its outputs very carefully. 

One thing seems clear. If the reports are accurate and of equal or better quality than those created by humans, AI bots such as DBOT is likely to replace junior analysts (or train them) and make experiences analysts more productive by doing the heavy lifting for them.

\section{Critique of DBOT}
\noindent
How does DBOT stack up against Damodaran?

\noindent
Damodaran offered the following summary remark on DBOT’s report on BYD:

\noindent
\textit{This is, for the most part, well done. If I were grading this write-up, I would suggest cutting down on the verbosity, since much of that write-up could have been condensed into half the pages. I have found that AI, because of its access to data and past write-ups, tends to overdo write-ups.}

The verbosity is addressable by instructing the critic agent accordingly, so that is an easy fix, but what about its substance? We pressed Damodaran for what he thought was missing from the analysis. He listed three questions, which would have framed his analysis:

\begin{enumerate}
    \item Will the transition to electric cars be as smooth as predicted?\\

\noindent
Damodaran’s prediction is that EV adoption will be slower than projected, with hybrids making a comeback. Arguably, the current infrastructure for gasoline and electric vehicles across the world is still tipped heavily towards the former, so perhaps hybrids will be more practical in the near future, and EV adoption will be slower than DBOT’s projections. This implies that revenue growth might be lower than what DBOT assumes. \\

    \item Where is the Chinese government in the BYD story? In every major Chinese company, Beijing is a key player and can make the difference.\\

\noindent
An earlier DBOT report did in fact refer to the role of the Chinese government, but it didn’t develop the theme and tell us why and how Beijing would support or thwart BYD’s global ambitions. Clearly Beijing is important for every major Chinese company.\\

\item Will the electric car market split into mass market and more upscale segments, with BYD dominating the former and TSLA the latter?\\

\noindent
This is a great question because if BYD were limited to the low end of the market, it would put pressure on margins, and hence revenues and valuation. 
\end{enumerate}

We refer to these as “framing questions,” in the sense that they drive the rest of the analysis. They seem obvious in retrospect, and yet they are anything but obvious in the moment. Herbert Simon describes such problems as ``ill-structured problems''\cite{simon1973structure}, whose structure lacks definition at the outset of the exercise, but which become defined by asking the right questions or making appropriate assumptions that constrain how an open-ended problem is structured for analysis.

For an illustration of other framing question, consider Damodaran's valuation of Nvidia in June 2023, in which he begins the analysis by asking ``Is AI an incremental or disruptive technology?''  This is an important question because disruptions tend to create new markets which can be big, but with high uncertainty about the market size and margins, whereas incremental technologies tend to create smaller markets and more certain margins. Damodaran says he previously viewed AI as an incremental technology, but changed his mind with the emergence of ChatGPT, which everyone could incorporate into their daily lives.

His follow-on question asks whether disruptions have been good or bad for investors in general, meaning, do they bolster valuations in general or lower them? Examples of disruptive changes over the last four decades are personal computers, the Internet, smartphones, and social media. He shows that these four tech disruptions have been beneficial to the broader market on average. 

Equally significantly, he asks about the distribution of winners and losers among suppliers of the new disruptive technology. This is important because it tells us the probability of a company succeeding or failing. He notes that disruptions have led to very few big winners but lots of hyped-up eventual losers looking to ride the disruptive bandwagon. Because of its size and positioning, Nvidia is likely to be one of the big winners. Because of its low risk of failure, its cost of capital is likely to be favorable relative to riskier companies.

DBOT is not currently able to generate a comparable set of framing questions on its own. In contrast, Damodaran’s questions are based on tacit knowledge that has been developed over decades of thinking about valuation, and keeping score of his mistakes and reasons for them. Damodaran says that his students also have trouble articulating these kinds of framing questions. They fail to look at the company with the appropriate big picture perspective. 

To what extent Damodaran’s tacit knowledge is learnable by DBOT from his writings is an open question at the moment. The political scientist Philip Tetlock asserts that a select few human ``superforecasters'' embody certain skills of the sort we see in Damodaran. 

Superforecasters tend to start with an ``outside view'' of the problem, such as, what would someone ask if they knew nothing about the domain under consideration? This prevents bias from seeping in early into the analysis. For example, we might ask about the probability of a recession next year. Most people would dive right into the factors that lead to recessions, such as the latest employment numbers, inflation, the Fed’s posture, along with a laundry list of other indicators about the state of the economy. Such a mindset immediately biases the analysis by grounding it in some chosen piece of data. In contrast, an outsider who knows nothing about economics or business would ask a more general question, such as ``how frequently have recessions occurred in the past hundred years?'' The answer to such a question provides the “base rate” of the phenomenon, namely, how likely is it to occur in general. The correct base rate grounds the analysis in the right ballpark, from where it can be nudged up or down depending on the specifics of the current context. Superforecasters pay close attention to base rates in making predictions. In contrast, people who dive into the details are relatively ungrounded and biased by their initial details.

Damodaran conjures up very similar types of grounding questions in his analysis of BYD and Nvidia to estimate the relevant base rates. For example, he asked whether disruptions are positive or negative for investors in the first place. His question about the role of Beijing for BYD is similar: how often is Beijing involved in a large Chinese company? The answer: almost always, so it makes sense to ask the question and dig deeper into it.

Perhaps the larger question is how we can turn DBOT into a superforecaster along the lines of Damodaran. We may need to get it to learn the properties that superforecasters display when confronted with complex open-ended problems. This type of reasoning may not yet be internalized by LLMs given the relative rarity of such analyses.

\section{Implications for Practice}

The paradigm shift in AI towards General Intelligence also signals a paradigm shift in the world of finance. The general intelligence paradigm provides a layer of intelligence on which intelligent agents can be easily designed to perform a variety of complex tasks that require judgment. AI machines like DBOT have significant implications for the practice of finance from the perspective of analysts, investors, and regulators.

As LLMs continue to evolve and improve, DBOT will benefit from these improvements. Specifically, longer contexts as well as fine-tuning to increase financial abilities will be very helpful to the subagents within DBOT. We have tested newer LLMs such as GPT-4.5, Gemini-2, Claude 3.7 Sonnet and have noticed improvements in their reasoning capabilities. Similarly, specialized LLMs such as BloombergGPT, FinTral, and FinGPT will improve certain subagents which need to reason over financial statements or financial news. The News agent can also be improved with DeepResearch-like capabilities in order to more accurately synthesize information.

Choosing the LLMs which compose subagents of DBOT is an interesting challenge.  Financial benchmarks, such as FinVis~\cite{wang2023finvis} and XFinBench~\cite{zhang2024finbench}, will help to improve our choice of models for DBOT. An interesting direction of future work is to compare improvements on benchmarks to subagent performance.

For analysts, if DBOT's capability rivals the average analyst and will only get better, it suggests that fewer analysts will be required. The question is, who will it replace, and who will it make better?

In the near-term, DBOT will make analysts a lot more productive, and it is likely to be of greater value to those who can ask the right kinds of framing questions. Indeed, for the foreseeable future, we would expect DBOT to be used with the “human in the loop,” where an expert casts the framing questions for DBOT. In the longer run, we should expect DBOT to get better at coming up with the framing questions. As we described in the last section, the work of \citet{tetlock2016superforecasting} on superforecasters provides clues on how the machine could learn how to come up with such questions.

For investors, DBOT democratizes long-term investing by providing an investment algorithm that is based on a set of investment principles of an authority whose writings are all in the public domain. There is currently no such product in the market, nor a comparable valuation expert with Damodaran's volume of training data and a well-defined methodology.

An additional benefit for investors is the AI’s transparent basis for its decisions, which are stated in its written report. The decisions are also amenable to scientific inquiry. For example, the value-oriented bias of the DBOT can be measured and related to other factors. In other words, by can measure the extent to which the Damodaran bot incorporates his systematic bias towards value investing, and seeing how its actual behavior “loads” on the various factors. DBOT can also be combined with other factors to create backtestable investment strategies tailored to investor preferences. 

Autonomous AI agents such as DBOT create new challenges for regulators, especially as such machines become available to retail investors. The investments are not made according to an explicit formula, as is the case with existing retail products such as ETFs. Rather, decisions are made by agents whose inner workings and reasoning are not fully understandable, even to their creators. Do retail investors need any protections when using such machines? Can reports by seemingly expert reports be biased to manipulate retail investment decisions? We indeed pushed DBOT to create reports both bullish and bearish on BYD in which it is able to create different convincing narratives, related to the same set of numbers. Retail investors may lack the ability to do the link the numbers and narrative properly and place too much trust in the AI. 

In general, accountability must be specified for autonomous AI investment managers. Who is accountable when the AI makes decisions that lose a lot of money? What is it engages in illegal activity such as acquiring inside information without anyone’s realization or deliberately moving the market in its desired direction? What if individuals blindly start to use DBOT through their own automated AI agents that are triggered by earnings call reports or the release of  major news? Would this create new kinds of systemic risk that cause major market distortions and flash crashes? 

Is it possible to sue the AI, just like one might file a lawsuit against a portfolio manager or an institution?

These are important questions that are not addressed within the current regulatory landscape which gives agency to humans and business entities, but doesn’t consider intelligent machines as having similar agency. In the era of AI, regulators may need to consider a new class of autonomous AI investment entities that are regulated in the same way as humans or corporate entities.

\bibliographystyle{apalike}
\bibliography{ref}

\begin{thebibliography}{}

\bibitem[Asness and Frazzini, 2013]{asness2013devil}
Asness, C. and Frazzini, A. (2013).
\newblock The devil in hml’s details.
\newblock {\em Journal of Portfolio Management}, 39(4):49--68.

\bibitem[Bhatia et~al., 2024]{bhatia-etal-2024-fintral}
Bhatia, G., Nagoudi, E. M.~B., Cavusoglu, H., and Abdul-Mageed, M. (2024).
\newblock {F}in{T}ral: A family of {GPT}-4 level multimodal financial large language models.
\newblock In Ku, L.-W., Martins, A., and Srikumar, V., editors, {\em Findings of the Association for Computational Linguistics: ACL 2024}, pages 13064--13087, Bangkok, Thailand. Association for Computational Linguistics.

\bibitem[Carhart, 1997]{carhart1997persistence}
Carhart, M.~M. (1997).
\newblock On persistence in mutual fund performance.
\newblock {\em The Journal of finance}, 52(1):57--82.

\bibitem[Chamberlain and Rothschild, 1982]{chamberlain1982arbitrage}
Chamberlain, G. and Rothschild, M. (1982).
\newblock Arbitrage, factor structure, and mean-variance analysis on large asset markets.

\bibitem[De~Prado, 2018]{de2018advances}
De~Prado, M.~L. (2018).
\newblock {\em Advances in financial machine learning}.
\newblock John Wiley \& Sons.

\bibitem[Dhar, 2024]{dhar2024paradigm}
Dhar, V. (2024).
\newblock The paradigm shifts in artificial intelligence.
\newblock {\em Commun. ACM}, 67(11):50–59.

\bibitem[Dhar et~al., 2000]{dhar2000discovering}
Dhar, V., Chou, D., and Provost, F. (2000).
\newblock Discovering interesting patterns for investment decision making with glower —a genetic learner overlaid with entropy reduction.
\newblock {\em Data Mining and Knowledge Discovery}, 4:251--280.

\bibitem[Fairhurst and Greene, 2025]{fairhurst2025how}
Fairhurst, D.~D. and Greene, D. (2025).
\newblock How much does chatgpt know about finance?
\newblock {\em Financial Analysts Journal}, 81(1):12--32.

\bibitem[Fama and French, 1993]{fama1993common}
Fama, E.~F. and French, K.~R. (1993).
\newblock Common risk factors in the returns on stocks and bonds.
\newblock {\em Journal of financial economics}, 33(1):3--56.

\bibitem[Giglio et~al., 2022]{giglio2022factor}
Giglio, S., Kelly, B., and Xiu, D. (2022).
\newblock Factor models, machine learning, and asset pricing.
\newblock {\em Annual Review of Financial Economics}, 14(1):337--368.

\bibitem[{Google}, 2025]{GoogleGeminiDeepResearch}
{Google} (2025).
\newblock {Gemini Overview: Deep Research}.
\newblock \url{https://gemini.google/overview/deep-research}.
\newblock Accessed: 2025-04-03.

\bibitem[Graham and Dodd, 1934]{graham1934security}
Graham, B. and Dodd, D. (1934).
\newblock {\em Security analysis}.
\newblock McGraw Hill.

\bibitem[Grinold and Kahn, 2000]{grinold2000active}
Grinold, R.~C. and Kahn, R.~N. (2000).
\newblock Active portfolio management.

\bibitem[Harvey et~al., 2016]{harvey2016and}
Harvey, C.~R., Liu, Y., and Zhu, H. (2016).
\newblock … and the cross-section of expected returns.
\newblock {\em The Review of Financial Studies}, 29(1):5--68.

\bibitem[Hong et~al., 2024]{hong2024metagpt}
Hong, S., Zhuge, M., Chen, J., Zheng, X., Cheng, Y., Wang, J., Zhang, C., Wang, Z., Yau, S. K.~S., Lin, Z., Zhou, L., Ran, C., Xiao, L., Wu, C., and Schmidhuber, J. (2024).
\newblock Meta{GPT}: Meta programming for a multi-agent collaborative framework.
\newblock In {\em The Twelfth International Conference on Learning Representations}.

\bibitem[Jegadeesh and Titman, 1993]{jegadeesh1993returns}
Jegadeesh, N. and Titman, S. (1993).
\newblock Returns to buying winners and selling losers: Implications for stock market efficiency.
\newblock {\em The Journal of finance}, 48(1):65--91.

\bibitem[Kelly et~al., 2019]{kelly2019characteristics}
Kelly, B.~T., Pruitt, S., and Su, Y. (2019).
\newblock Characteristics are covariances: A unified model of risk and return.
\newblock {\em Journal of Financial Economics}, 134(3):501--524.

\bibitem[Lu et~al., 2024]{lu2024ai}
Lu, C., Lu, C., Lange, R.~T., Foerster, J., Clune, J., and Ha, D. (2024).
\newblock The ai scientist: Towards fully automated open-ended scientific discovery.
\newblock {\em arXiv preprint arXiv:2408.06292}.

\bibitem[{OpenAI}, 2025]{openai2025deepresearch}
{OpenAI} (2025).
\newblock Introducing deep research.
\newblock \url{https://openai.com/index/introducing-deep-research/}.
\newblock Accessed: 2025-04-03.

\bibitem[Samuel, 1959]{samuel1959some}
Samuel, A.~L. (1959).
\newblock Some studies in machine learning using the game of checkers.
\newblock {\em IBM Journal of research and development}, 3(3):210--229.

\bibitem[Sharpe, 1964]{sharpe1964capital}
Sharpe, W.~F. (1964).
\newblock Capital asset prices: A theory of market equilibrium under conditions of risk.
\newblock {\em The journal of finance}, 19(3):425--442.

\bibitem[Simon, 1973]{simon1973structure}
Simon, H.~A. (1973).
\newblock The structure of ill structured problems.
\newblock {\em Artificial intelligence}, 4(3-4):181--201.

\bibitem[Steiner, 2012]{steiner2012automate}
Steiner, C. (2012).
\newblock {\em Automate This: How Algorithms Came to Rule Our World}.

\bibitem[Tetlock and Gardner, 2016]{tetlock2016superforecasting}
Tetlock, P.~E. and Gardner, D. (2016).
\newblock {\em Superforecasting: The art and science of prediction}.
\newblock Random House.

\bibitem[Wang et~al., 2023]{wang2023finvis}
Wang, Z., Li, Y., Wu, J., Soon, J., and Zhang, X. (2023).
\newblock Finvis-gpt: A multimodal large language model for financial chart analysis.
\newblock {\em arXiv preprint arXiv:2308.01430}.

\bibitem[Wei et~al., 2022]{wei2022chain}
Wei, J., Wang, X., Schuurmans, D., Bosma, M., Xia, F., Chi, E., Le, Q.~V., Zhou, D., et~al. (2022).
\newblock Chain-of-thought prompting elicits reasoning in large language models.
\newblock {\em Advances in neural information processing systems}, 35:24824--24837.

\bibitem[Weiss, 2006]{weiss2006after}
Weiss, D.~M. (2006).
\newblock {\em After the Trade Is Made, Revised Ed.: Processing Securities Transactions}.
\newblock Penguin.

\bibitem[Wu et~al., 2023]{wu2023bloomberggpt}
Wu, S., Irsoy, O., Lu, S., Dabravolski, V., Dredze, M., Gehrmann, S., Kambadur, P., Rosenberg, D., and Mann, G. (2023).
\newblock Bloomberggpt: A large language model for finance.
\newblock {\em arXiv preprint arXiv:2303.17564}.

\bibitem[Xie et~al., 2023]{xie2023pixiu}
Xie, Q., Han, W., Zhang, X., Lai, Y., Peng, M., Lopez-Lira, A., and Huang, J. (2023).
\newblock Pixiu: A large language model, instruction data and evaluation benchmark for finance.
\newblock {\em arXiv preprint arXiv:2306.05443}.

\bibitem[Zhang et~al., 2023]{zhang2023instruct}
Zhang, B., Yang, H., and Liu, X.-Y. (2023).
\newblock Instruct-fingpt: Financial sentiment analysis by instruction tuning of general-purpose large language models.
\newblock {\em arXiv preprint arXiv:2306.12659}.

\bibitem[Zhang et~al., 2024]{zhang2024finbench}
Zhang, Z., Cao, Y., and Liao, L. (2024).
\newblock Finbench: Benchmarking {LLM}s in complex financial problem solving and reasoning.

\bibitem[Zhou et~al., 2024]{zhou2024finrobot}
Zhou, T., Wang, P., Wu, Y., and Yang, H. (2024).
\newblock Finrobot: Ai agent for equity research and valuation with large language models.
\newblock {\em arXiv preprint arXiv:2411.08804}.

\end{thebibliography}

\clearpage

\appendix
\section{DBOT Algorithm}
\label{app:algorithm}

\begin{algorithm}[ht!]
\caption{Valuation and Report Generation Algorithm}

\SetKwInOut{Require}{Require}
\SetKwInOut{Ensure}{Ensure}

\Require{
  \begin{itemize}
    \item $C_n$: Company name
    \item $C_t$: Ticker symbol
    \item $D_C$: Data related to $C_n$ (e.g., income statements, analyst consensus, industry comparables)
    \item $v$: Valuation (e.g., from a free cash flow or Damodaran-based model)
    \item $I$: Spreadsheet inputs (e.g., income statements, value drivers)
    \item $f^{fcf}$: Free cash flow valuation function,
      $  f^{fcf}: I \,\mapsto\, v$
    \item $s$: News search function,
$        s: C_t \,\mapsto\, \text{text} $
    \item $f^{llm}$: Large language model function,
       $ f^{llm}: \text{text} \,\mapsto\, \text{text},$
      with specialized sub-functions:
      \begin{itemize}
        \item $f^{vd} = f^{llm}(P \;\|\; D_C \;\mapsto\; I)$: Maps a prompt $P$ and $D_C$ to inputs $I$.
        \item $f^{report} = f^{llm}(P \;\|\; I \;\|\; v \;\mapsto\; R)$: Maps $P$, $I$, and $v$ to a textual report $R$.
        \item $f^{D}_{news} = f^{llm}(P \;\|\; I \;\|\; v \;\|\; \text{news} \,\mapsto\, I)$: Incorporates news into $I$.
        \item $f^{valuation} = f^{llm}(P \;\|\; I \,\mapsto\, I)$: Produces updated inputs $I$ for valuation.
        \item $f^{news\_summary} = f^{llm}(P \;\|\; \text{news} \,\mapsto\, \text{text})$: Summarizes news textually.
        \item $f^{plot} = \texttt{Pyth}\bigl(f^{llm}(P \;\|\; I \,\mapsto\, \text{code})\bigr) \,\mapsto\, \text{image}$:
              Uses an LLM-generated Python code to produce a plot/image.
      \end{itemize}
  \end{itemize}
}
\small

\textbf{Initialize:}\\
\Indp
$P_{D_C \to I}$: Prompt for mapping $D_C$ to $I$.\\
\Indm
\BlankLine

\textbf{Initial Valuation Setup:} $f^{vd}(I) = v$.

\Begin{
  \textbf{Initial Waterfall Valuation}{
    \tcp{Each sub-step updates $I$ via $f^{valuation}$ and then applies $f^{fcf}$ to compute $v$.}
    $v \gets f^{fcf}\bigl(f^{valuation}(P_{market} \;\|\; I)\bigr)$ \tcp*[r]{Market valuation}
    $v \gets f^{fcf}\bigl(f^{valuation}(P_{sensitivity} \;\|\; I)\bigr)$ \tcp*[r]{Sensitivity analysis}
    $v \gets f^{fcf}\bigl(f^{valuation}(P_{consensus} \;\|\; I)\bigr)$ \tcp*[r]{Consensus analysis}
    $v \gets f^{fcf}\bigl(f^{valuation}(P_{comparables} \;\|\; I)\bigr)$ \tcp*[r]{Comparable analysis}
    $v \gets f^{fcf}\bigl(f^{D}_{news}\bigl(s(C_t)\bigr) \bigr)$ \tcp*[r]{Incorporate news into $I$, then get $v$}

    \textbf{Initialize } $P_{\text{router}}$: Prompt to guide the LLM's choice of sub-procedures.\;
    \While{\text{not converged}}{
      $\text{route}, \text{instruction} \gets f^{llm}(P_{\text{router}} \;\|\; I \;\|\; v)$ \tcp*[r]{LLM decides which step to run next}
      \uIf{$\text{route} = \text{"market"}$}{
        $v \gets f^{fcf}\bigl(f^{valuation}(P_{market} \;\|\; I)\bigr)$
      }
      \uElseIf{$\text{route} = \text{"sensitivity"}$}{
        $v \gets f^{fcf}\bigl(f^{valuation}(P_{sensitivity} \;\|\; I)\bigr)$
      }
      \uElseIf{$\text{route} = \text{"consensus"}$}{
        $v \gets f^{fcf}\bigl(f^{valuation}(P_{consensus} \;\|\; I)\bigr)$
      }
      \uElseIf{$\text{route} = \text{"comparables"}$}{
        $v \gets f^{fcf}\bigl(f^{valuation}(P_{comparables} \;\|\; I)\bigr)$
      }
      \uElseIf{$\text{route} = \text{"news"}$}{
        $v \gets f^{fcf}\bigl(f^{D}_{news}\bigl(s(C_t)\bigr)\bigr)$
      }
      \Else{
        \textbf{break} \tcp*[r]{If LLM chooses to end or an unknown route is given}
      }
      \tcp{Optionally update $I$ if needed, e.g. $I \leftarrow \dots$}
    }
  }

  \BlankLine

  \textbf{Valuation and Report Generation}{
    \tcp{1) Convert raw data $D_C$ into inputs $I$.}
    $I \gets f^{llm}(P_{D_C \to I} \;\|\; D_C)$ \tcp*[r]{Generate spreadsheet inputs from $D_C$}

    \tcp{2) Perform a direct FCF-based valuation.}
    $v \gets f^{fcf}(I)$ \tcp*[r]{Compute valuation}

    \tcp{3) Define a prompt for generating the final report.}
    $P_{I \,\|\; v \to R}$: Prompt for producing a textual report from $I$ and $v$.

    \tcp{4) Generate the report.}
    $R \gets f^{llm}(P_{I \,\|\; v \to R} \;\|\; I \;\|\; v)$ \tcp*[r]{Produce textual report $R$}
  }

}

\end{algorithm}

\clearpage
\section{BYD Report}
\label{app:byd-report}


\section*{Supplementary material}

**Appendix B. BYD Report**

# BYD in 2024: Riding the EV Wave or Struggling Through the Competitive Storm?

## 1. Review Historical Performance

BYD has demonstrated remarkable financial growth and operational resilience over the past few years. Revenue has surged from CNY 143.0 billion in 2019 to an estimated CNY 662.6 billion in 2023, underpinned by a robust CapEx strategy focused on expanding production capacity and innovating their product line. Notably, BYD's revenue grew by 24% year-over-year in Q3 2024, outpacing competitors like Tesla and Li Auto. This surge underscores the firm's successful capacity scaling and enhanced market penetration in an increasingly saturated market.

The historical performance of BYD reflects not just growth in numbers but also a consistent focus on long-term strategic goals. One of the core drivers of this financial uptrend has been BYD's deep investment in infrastructure, research, and development, which allowed the company to diversify its offerings in the electric vehicle (EV) segment and cater to a wide spectrum of customers. In addition to increasing revenue, BYD's operational efficiency has led to consistent growth in its market share, with significant improvements in production technology enabling faster and more cost-effective output. This strategic approach has been pivotal for BYD to stay ahead in a highly competitive EV market that is dominated by well-established giants like Tesla and newer but aggressive players such as NIO and Li Auto.

Operating margins have also trended upward, improving from approximately 2.97% in 2020 to 5.88% in 2023, suggesting effective cost management and strategic investments. The fact that BYD managed to maintain margin improvements even as the competitive landscape became more challenging speaks to the company's strength in optimizing both manufacturing and operational expenses. Consistent margin improvements confirm BYD's competitiveness, despite significant industry pressures from rising raw material costs and fluctuating global demand. Record-breaking monthly deliveries, as shown in **Figure 1: BYD Monthly NEV Deliveries Over Time**, further accentuate the company's operational efficiency, highlighting strong market demand and production capabilities.

**Figure 1: BYD Monthly NEV Deliveries Over Time**

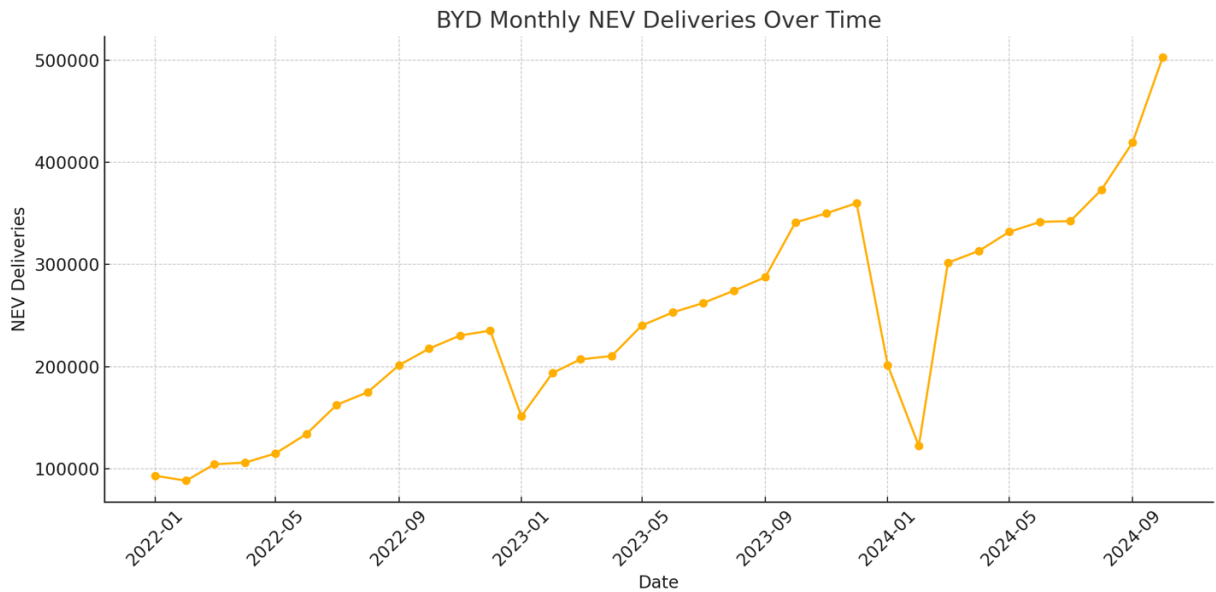

These delivery trends represent BYD's growing ability to meet consumer demand while addressing supply chain challenges. The steady increase in monthly deliveries over time illustrates how BYD has adeptly managed its supply chain, enhanced production capacity, and improved vehicle offerings. Moreover, the growth in NEV (New Energy Vehicle) deliveries positions BYD as a leader in the transition to sustainable transportation.

**2. Forecast Future Performance**

Future projections for BYD consider both its historical financial trajectory and the evolving market environment. Revenue growth is projected at 10% for 2024, driven by market expansion initiatives and a solid competitive position. From 2024-2028, a compound annual growth rate (CAGR) of 7% is expected, reflecting both the opportunities and the headwinds BYD faces, such as intensified competition and potential tariff impacts.

One of the main opportunities for BYD moving forward is the expanding consumer base for electric vehicles globally. Governments around the world are increasingly adopting policies to encourage EV adoption, which provides BYD with new market entry opportunities. The company's focus on producing affordable and reliable electric vehicles makes it uniquely positioned to capture this burgeoning demand, particularly in developing markets where affordability is key. Moreover, BYD's emphasis on a diversified product line, from compact urban vehicles to luxury EVs, ensures that it caters to a wide variety of customer needs.

Operating margins are projected to stabilize around 7% by 2028, benefiting from economies of scale and ongoing cost optimizations, particularly through European manufacturing investments. The ongoing expansion into the European market, which includes establishing new production facilities, is crucial for achieving these projected margins. European operations are expected to bring about efficiencies in logistics and production, thereby reducing costs and enhancing BYD's competitiveness in this high-value market. The valuation models indicate a target pre-tax operating margin of 7.75%, and achieving this goal will require BYD to focus on technological innovations that can streamline manufacturing processes, reduce costs, and improve vehicle performance.

Sales to capital ratios, key efficiency metrics, are expected to start at 1.2 and improve to 1.6 in subsequent years, signaling better capital utilization as BYD's infrastructure matures and market strategies solidify. Effective capital deployment is integral to maintaining growth momentum. This improved ratio is indicative of the company's evolving strategy of not just expanding physical production capabilities, but also investing in technology that maximizes returns on those investments.

The chart **Figure 2: TEV/EBITDA Multiples of BYD and Competitors** provides additional insight into BYD's relative valuation compared to its peers. With a TEV/EBITDA multiple of 8.9x, BYD is valued below the industry average of 9.84x, suggesting potential upside compared to competitors like Tesla and NIO.

**Figure 2: TEV/EBITDA Multiples of BYD and Competitors**

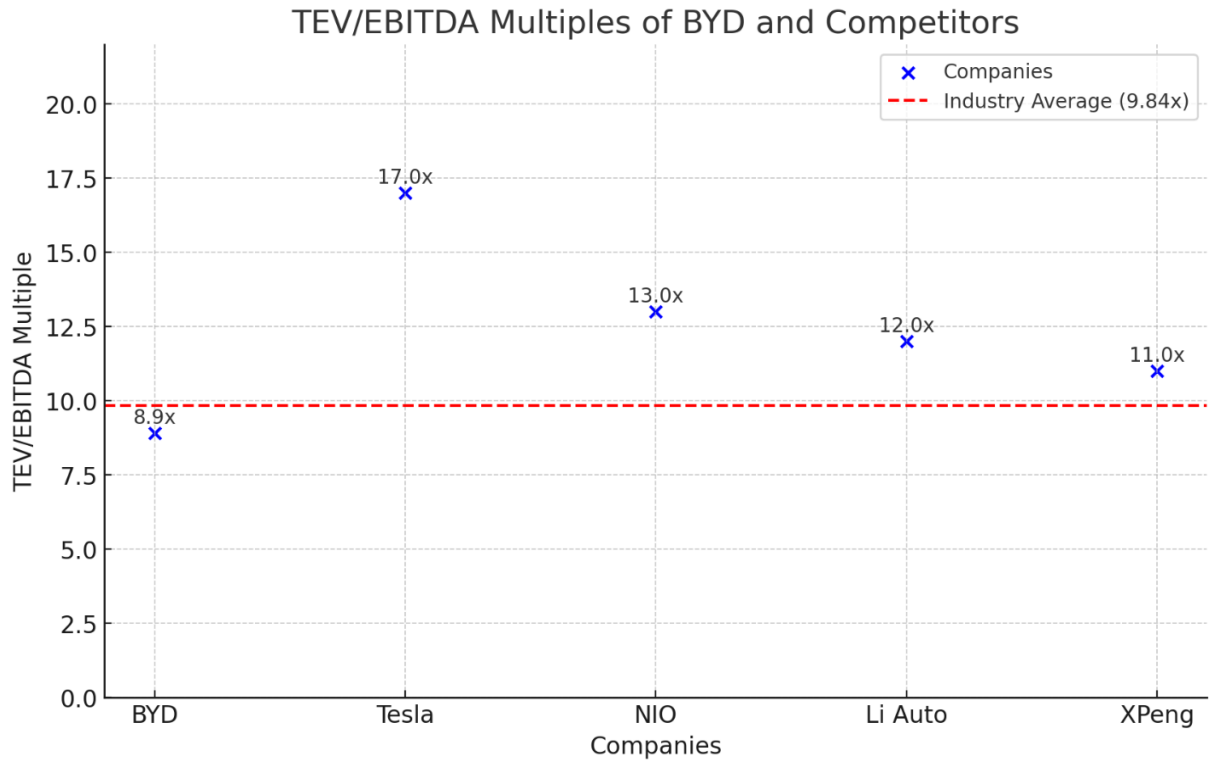

The TEV/EBITDA multiple analysis emphasizes BYD's potential as an undervalued player within the industry. Compared to Tesla, whose multiple is considerably higher, BYD offers a more balanced risk-reward profile for investors. Tesla's higher multiple may reflect its strong brand and higher growth expectations, but BYD's lower valuation suggests more room for appreciation as the company continues to strengthen its financial metrics and global presence.

### 3. Sensitivity Analysis

Sensitivity analysis reveals strong resilience in BYD's valuation under different scenarios. Key assumptions include:

- Gross revenue growth ranging between 5%-12% annually.
- Operating margins fluctuating between 6%-8%.

**BYD Valuation Sensitivity Analysis**

| Revenue Growth \ Operating Margin | 6% | 6.5% | 7% | 7.5% | 8% |
| --- | --- | --- | --- | --- | --- |
| 5% | 417.54 | 424.02 | 430.51 | 436.99 | 443.47 |
| 6% | 420.87 | 427.41 | 433.96 | 440.50 | 447.05 |
| 7% | 424.20 | 430.80 | 437.41 | 444.02 | 450.63 |
| 8% | 427.53 | 434.19 | 440.86 | 447.53 | 454.20 |
| 9% | 430.85 | 437.59 | 444.32 | 451.05 | 457.78 |
| 10% | 434.18 | 440.98 | 447.77 | 454.56 | 461.35 |
| 11% | 437.51 | 444.37 | 451.22 | 458.07 | 464.93 |
| 12% | 440.84 | 447.76 | 454.67 | 461.59 | 468.50 |

The sensitivity analysis is particularly useful in understanding how different market dynamics could impact BYD's future performance. Even under conservative estimates of 6% revenue growth and a 6% operating margin, BYD's valuation remains favorable compared to current market expectations. This resilience is largely attributed to BYD's diversified product portfolio, efficient production methods, and strong presence in both domestic and international markets. The strategic focus on cost control and market expansion buffers against downside risks.

BYD's ability to adapt to various scenarios also hinges on its deep understanding of the local markets it operates in, and its agility in modifying product lines or cost structures to meet changing consumer demands. As competition intensifies, maintaining profitability will require BYD to be proactive in managing costs, innovating on the product front, and finding new ways to appeal to customers in both established and emerging markets.

**4. Macro-Economic Factors**

Broader macro-economic factors are pivotal in assessing BYD's valuation. Interest rates at 4.37% and an initial cost of capital of 8.89% represent a cautious yet stable investment climate. Intense price competition within the EV sector, alongside geopolitical factors like EU tariffs and North American trade policies, will necessitate strategic agility from BYD.

The macro-economic environment also includes fluctuating currency exchange rates, raw material price volatility, and shifting trade dynamics, all of which could affect BYD's cost structure and profitability. For example, increased tariffs on Chinese imports into Europe could have a significant impact on BYD's bottom line. The company's investment in localized production in Europe is a strategic move to mitigate this risk, ensuring that it can still be competitive even in the face of challenging trade policies.

The chart **Figure 3: Global Electric-Car Revolution Set to Take Off** highlights the projected growth in global EV adoption, with China leading the market. This growth trend presents significant opportunities for BYD to expand internationally, particularly in Europe and other emerging markets.

**Figure 3: Global Electric-Car Revolution Set to Take Off**

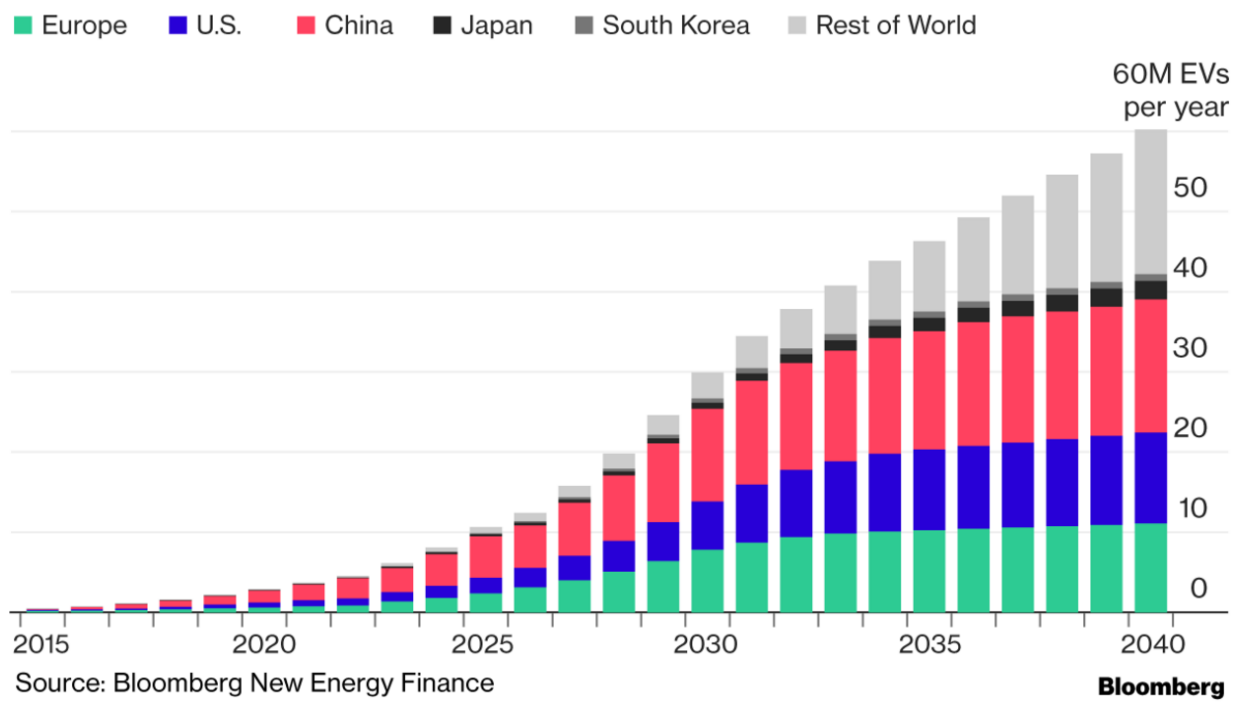

The global EV growth trajectory, as illustrated in Figure 3, signifies that the demand for electric vehicles will continue to grow across all regions, with China being at the forefront. This is driven by strong policy support, subsidies, and rapid advancements in EV technologies within China. For BYD, which already holds a leading market position in China, leveraging this growth to expand into new regions will be critical. In particular, BYD's focus on cost-effective and highly efficient vehicles will resonate well with new customers looking to make the switch to electric mobility.

BYD's targeted investments in European manufacturing are a key component of its strategy to counteract tariff impacts and capitalize on local advantages. These investments, coupled with ongoing strong performance despite both internal and external pressures, position BYD favorably for future growth amidst a challenging global EV landscape. Europe, as a market, is critical for BYD due to its stringent emission regulations and growing consumer demand for greener alternatives. By investing in localized production, BYD not only reduces its exposure to trade-related risks but also aligns itself with regional consumer preferences, thus strengthening its market position.

**5. Valuation Analysis**

The valuation of BYD has been meticulously modeled, considering both the projected revenues and operating income over the next decade. The table below presents the estimated valuation inputs, taking into account expected revenue growth rates, EBIT margins, and reinvestment needs.

| **Base year** | **Next year** | **Years 2-5** | **Years 6-10** | **After year 10** |
|---|---|---|---|---|
| Revenues (a) | $682,291.83 | 10.0% | 7.00% | Changes to 4.37% |
| Operating margin (b) | 5.88% | 5.0% | Moves to 6.67% | 7.00% |
| Tax rate | 17.07% | | 17.07% | Changes to 25.00% |
| Sales to Capital (c) | | 1.20 | 1.20 | 1.60 |
| Return on capital | 35.39% | 7.94% | | 8.70% |
| Cost of capital (d) | | 8.89% | | 8.70% |

These valuation inputs are crucial to understanding BYD's trajectory and tie directly into the broader narrative of the company's growth potential and market strategy. The revenue growth rate is expected to start strong at 10% next year, supported by continued expansion and penetration into international markets. However, over time, as the market matures and competition intensifies, the growth rate gradually declines to 4.37%, indicating a shift from aggressive expansion to more stable and sustainable growth.

The operating margin is expected to improve from 5.88% to 7.00% after Year 10, showcasing BYD's success in achieving greater economies of scale and operational efficiency through strategic investments in technology and localized production. This links to the story of cost optimization efforts and geographical expansion, which help mitigate pressures such as rising costs and geopolitical risks.

Tax rate changes from 17.07% to 25.00% after Year 10, reflecting the evolving tax landscape as BYD increases its presence in more developed markets where tax rates tend to be higher. The rise in tax obligations is an important consideration for BYD, especially as it aims to achieve profitability while expanding its global footprint.

The Sales to Capital ratio is projected to improve to 1.60 after Year 10, reflecting better capital utilization as a result of BYD's efficient reinvestment strategies. By focusing on expanding capacity while ensuring that capital investments yield optimal returns, BYD is poised to maintain a healthy growth trajectory. This aligns with the broader story of BYD maturing as a global automaker that successfully balances growth and capital efficiency.

The return on invested capital (ROIC) is initially high at 35.39% but is expected to normalize to a marginal ROIC of around 8.70%. This pattern is typical for companies transitioning from high-growth phases to more stable growth phases. The declining ROIC is balanced by a lowering cost of capital from 8.89% to 8.70%, indicating improving risk management and reduced cost of financing as BYD establishes itself firmly in global markets.

The terminal value was estimated at $806,707.13 million, discounted to its present value of $346,027.61 million. With a cumulative present value of cash flows over the next decade at $41,092.37 million, the total estimated value of operating assets comes to $387,119.98 million. After adjusting for debt of $46,886.03 million and adding non-operating assets, including cash holdings of $91,734.52 million, the estimated value of BYD's equity amounts to $461,817.08 million. With 1,098 shares outstanding, the estimated value per share stands at $420.60, significantly higher than the current trading price of $253.60, implying that the market undervalues BYD's equity at just 60.3% of its estimated intrinsic value.

**Strategic Investments and Market Expansion**

BYD's expansion into Europe through new manufacturing facilities is critical for its long-term growth. These investments aim to mitigate the impact of EU tariffs on Chinese EVs, enhancing BYD's position in the European market. BYD's focus on showcasing advanced EV models at international auto shows, despite geopolitical headwinds, underscores its commitment to innovation and market leadership.

Furthermore, BYD's strategic investments are not limited to infrastructure alone; they also encompass R&D for vehicle technology, battery efficiency, and autonomous driving capabilities. By pushing the boundaries of technology, BYD aims to offer vehicles that are not just affordable but also state-of-the-art in terms of features and performance. Efforts to localize production through partnerships and joint ventures reflect a forward-looking approach to globalization. This strategy not only reduces operational costs but also aligns BYD with regional regulations, facilitating smoother market entry and expansion. With sustained focus on technological innovation and cost management, BYD is well-positioned to capture significant market share in both developed and emerging markets.

The chart **Figure 4: China continues its strong EV momentum into H1 2023** provides insight into BYD's market leadership, with a 21% share of the Chinese EV market in H1 2023. This dominant position is critical for maintaining momentum as BYD expands its international footprint.

**Figure 4: China continues its strong EV momentum into H1 2023**

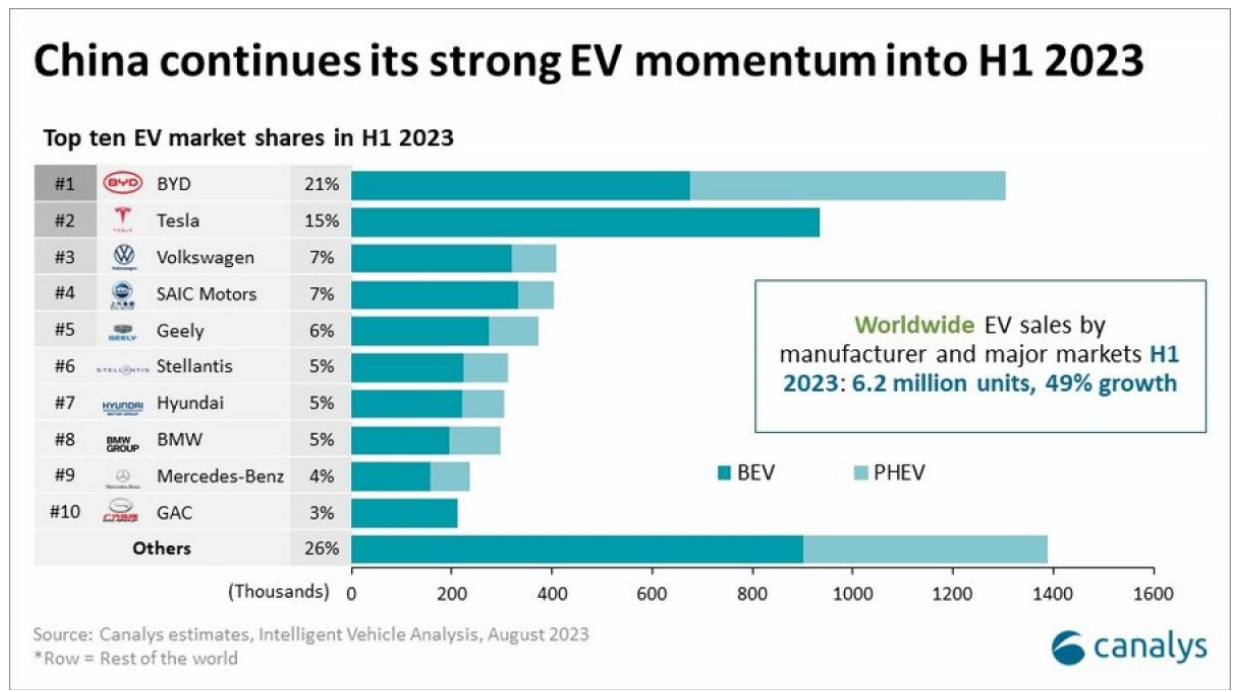

BYD's success in China, as highlighted in Figure 4, forms a strong foundation for its global ambitions. China remains the largest market for electric vehicles, and BYD's leadership there means it can leverage scale, brand recognition, and consumer trust to accelerate its expansion in other regions. Moreover, the company's product mix, which includes both battery electric vehicles (BEVs) and plug-in hybrid electric vehicles (PHEVs), allows it to cater to varied market requirements, further strengthening its market positioning.

## Competitive Landscape and Market Dynamics

However, BYD faces a fiercely competitive landscape, particularly in China, where price wars are putting pressure on margins. The push for supplier cost reductions across the industry could threaten profit stability. To maintain its edge, BYD will need to uphold stringent cost controls while continuing to innovate and deliver value to its customers.

The competition in the Chinese EV market is characterized by aggressive pricing strategies, with companies like NIO, XPeng, and others offering significant discounts to gain market share. This environment creates downward pressure on prices, impacting margins. Despite these challenges, BYD has managed to remain profitable by focusing on cost-efficient production processes and leveraging economies of scale. However, to maintain its competitive advantage, the company will need to continue to innovate not only in vehicle technology but also in its approach to cost management and supplier negotiations.

BYD's position in the global market also presents both opportunities and challenges. As other automakers ramp up their EV production capabilities, BYD must differentiate itself through superior technology, battery efficiency, and pricing strategies that make its vehicles attractive across different customer segments. The push towards autonomous driving and connected car technologies is another area where BYD can build an advantage, but it will require substantial investments in R&D and strategic partnerships.

The evolving regulatory landscape is another dynamic that BYD must navigate carefully. With different countries setting varying emissions standards and offering different levels of subsidies and incentives for electric vehicles, a one-size-fits-all strategy will not suffice. BYD's ability to adapt to each market's unique requirements—through localized production, tailored product offerings, and strategic alliances—will be key to its success in expanding its global footprint.



\end{document}